\def\year{2020}\relax
\documentclass[letterpaper]{article} 
\usepackage{aaai20}  
\usepackage{times}  
\usepackage{helvet} 
\usepackage{courier}  
\usepackage[hyphens]{url}  
\usepackage{graphicx} 
\usepackage{graphicx}  
\usepackage{amssymb}
\usepackage{pifont}
\newcommand{\cmark}{\ding{51}}%
\newcommand{\xmark}{\ding{55}}%
\usepackage{float}
\usepackage{booktabs}
\usepackage{threeparttable}
\usepackage[table]{xcolor}
\usepackage{tcolorbox}

\definecolor{Gray}{gray}{0.9}
\usepackage{color,soul}
\title{Evaluating Commonsense in Pre-trained Language Models}
\author{
Xuhui Zhou,\textsuperscript{\rm 1}\thanks{Work done while at Westlake University}
Yue Zhang,\textsuperscript{\rm 2}
Leyang Cui,\textsuperscript{\rm 2 \rm 3}
Dandan Huang\textsuperscript{\rm 2}\\
\textsuperscript{\rm 1}University of Washington\\
\textsuperscript{\rm 2}School of Engineering, Westlake
University\\
\textsuperscript{\rm 3}Zhejiang University\\
xuhuizh@uw.edu, yue.zhang, cuileyang, huangdandan@westlake.edu.cn
}
\begin{document}

\maketitle
\begin{abstract}
Contextualized representations trained over large raw text data have given remarkable improvements for NLP tasks including question answering and reading comprehension. There have been works showing that syntactic, semantic and word sense knowledge are contained in such representations, which explains why they benefit such tasks. However, relatively little work has been done investigating commonsense knowledge contained in contextualized representations, which is crucial for human question answering and reading comprehension. We study the commonsense ability of GPT, BERT, XLNet, and RoBERTa by testing them on seven challenging benchmarks, finding that language modeling and its variants are effective objectives for promoting models' commonsense ability while bi-directional context and larger training set are bonuses. We additionally find that current models do poorly on tasks require more necessary inference steps. Finally, we test the robustness of models by making dual test cases, which are correlated so that the correct prediction of one sample should lead to correct prediction of the other. Interestingly, the models show confusion on these test cases, which suggests that they learn commonsense at the surface rather than the deep level. We release a test set, named CATs publicly, for future research.
\end{abstract}

\section{Introduction}
Contextualized representations trained over large-scale text data have given remarkable improvements to a wide range of NLP tasks, including natural language inference \cite{Bowman2015ALA}, question answering \cite{rajpurkar-etal-2018-know} and reading comprehension \cite{Lai2017RACELR}. Giving new state-of-the-art results that approach or surpass human performance on several benchmark datasets, it is an interesting question what types of knowledge are learned in pre-trained contextualized representations in order to better understand how they benefit the NLP problems above. There has been work investigating the nature of syntactic \cite{liu-etal-2019-linguistic}, semantic \cite{liu-etal-2019-linguistic} and word sense \cite{kim-etal-2019-probing} knowledge contained in such contextualized representations, in particular BERT \cite{devlin-etal-2019-bert}, showing that such knowledge can be effectively learned via language model (LM) pre-training over large scale data.

Commonsense knowledge spans ``a huge portion of human experience, encompassing
knowledge about the spatial, physical, social, temporal, and
psychological aspects of typical everyday life. '' \cite{Liu2004ConceptNetA}. Intuitively, such knowledge is at least as useful as semantic and syntactic knowledge in natural language inference, reading comprehension and coreference resolution. For example, the word ``it'' in the sentence ``the dog cannot cross the street because it is too X'' can refer to three different entities when the word ``X'' is ``timid'', ``wide'' and ``dark'', respectively, and resolving such ambiguity can require that a system has relevant commonsense knowledge beyond the sentence level. However, relatively little work has been conducted on systematically evaluating the nature of commonsense knowledge learned in contextualized representations.

\begin{table*}[!htbp]
\centering
\begin{tabular}{cl c c}
\toprule
 & \hfill Token-level& \\
\hline
CA &They broadcast an announcement, \textbf{but} a subway came into the station and I couldn't hear it. & \cmark \\ 
 & They broadcast an announcement, \textbf{before} a subway came into the station and I couldn't hear it . & \xmark \\
\hline
WSC &The trophy doesn't fit into the brown suitcase because the \textbf{trophy} is too large. & \cmark \\
 & The trophy doesn't fit into the brown suitcase because the \textbf{suitcase} is too large. & \xmark \\
\hline
SM &  money can be used for buying \textbf{cars} & \cmark \\
& money can be used for buying \textbf{stars} & \xmark \\
\toprule
& \hfill Sentence-level &\\
\hline
SMR &  $\neg$`` he put an elephant into the fridge'' (because) $\leftarrow$ an elephant is much bigger than a fridge . & \cmark \\
& $\neg$`` he put an elephant into the fridge " (because) $\leftarrow$ elephants are usually gray... & \xmark \\
& $\neg$`` he put an elephant into the fridge " (because) $\leftarrow$ an elephant cannot eat a fridge .  & \xmark \\
\hline
SWAG &  Someone unlocks the door and they go in. $\rightarrow$ Someone leads the way in.& \cmark \\
& Someone unlocks the door and they go in. $\rightarrow$ Someone opens the door and walks out. &\xmark
\\
& Someone unlocks the door and they go in. $\rightarrow$ Someone walks out of the driveway. &\xmark\\
& Someone unlocks the door and they go in. $\rightarrow$ Someone walks next to someone and sits on a pew. &\xmark\\
\hline 
HellaSwag &  A carved pumpkin with a light in it glows on a counter. Supplies for carving are then shown. & \\
& $\rightarrow$ A woman cuts the top off the pumpkin, emptying the seeds. & \cmark \\
& $\rightarrow$ she cuts down all the pieces and dumps them in a trash bin in the end. &\xmark\\
& $\rightarrow$ she then carves the traced lines to cut out the design. &\xmark\\
& $\rightarrow$ she tapes the top shut as the continue carving the pumpkin. &\xmark\\
\hline
ARCT &  People can choose not to use Google $\wedge$ Other search engines don’t redirect to Google \\ & $\rightarrow$ Google is not a harmful monopoly & \cmark \\
&  People can choose not to use Google $\wedge$  All other search engines redirect to Google \\ & $\rightarrow$ Google is not a harmful monopoly & \xmark \\
\toprule

\end{tabular}
\caption{Example of reframed test instances corresponding to each of our test task. The key word is \textbf{bolded} in token-level tasks. $\wedge$, $\neg$, $\leftarrow$ and $\rightarrow$ are used for showing the logic flows and replaced by natural language in actual test data.}\smallskip
\label{table1}
\end{table*}

We fill this gap by evaluating five state-of-the-art contextualized embedding models on seven commonsense benchmarks. The models include off-the-shelf embeddings\footnote{https://github.com/huggingface/transformers} from GPT \cite{rad-2018}, GPT2 \cite{radford2019language}, BERT \cite{devlin-etal-2019-bert}, XLNet \cite{zhilin-19} and RoBERTa \cite{liu2019roberta}, and the benchmarks include Conjunction Acceptability, Sense Making \cite{wang-etal-2019-make}, Winograd Schema Challenge \cite{Levesque:2012:WSC:3031843.3031909}, SWAG \cite{zellers-etal-2018-swag}, HellaSwag \cite{zellers-etal-2019-hellaswag}, Sense Making with Reasoning \cite{wang-etal-2019-make}, and Argument Reasoning Comprehension \cite{habernal-etal-2018-argument}. We evaluate commonsense knowledge contained in the above models by unifying the form of all the datasets and comparing LM perplexities on positive and negative samples (i.e., sentences that make sense and those that do not make sense, respectively). Commonsense contained in our data covers a wide range of subjects, from physical world knowledge to social conventions, from scientific domains to daily life scenes. We further categorize them by the difficulty level, namely the number of inference steps necessary in making sense.

We reframe the datasets in order to conduct both word- and sentence-level testing. For word-level testing, negative samples are drawn by replacing words from positive samples. We are concerned about nouns, verbs, adjectives, adverbs, pronouns and conjunctions, which reflect different aspects of commonsense. For example, while verbs such as ``buy, throw, sell ...'' are relatively more associated with event knowledge, conjunctions such as ``because, but, so ...'' are more associated with logical reasoning. For sentence-level testing, negative examples are drawn by replacing a full subsentences (such as a clause) with irrelevant or conflicting contents. Sentence-level tests concern more about commonsense inference.

From the results we have four salient observations. First, the pre-trained models give consistently better performances than random baselines, which demonstrates that language model pre-training is useful for learning commonsense knowledge. Second, models based on bi-directional contexts such as BERT, XLNet and RoBERTa are stronger in learning commonsense knowledge compared to those based on uni-directional contexts, such as GPT and GPT2. Third, more commonsense knowledge can be learned from larger training sets, which conforms well to the intuition. Fourth, the models have a certain degree of commonsense reasoning ability. However, as the number of necessary inference steps increase, the model performances drop, which shows that commonsense is still a big challenge that is not completely solved by pre-trained contextualized language models (LMs).

Finally, we further test the robustness of the five models by making dual test samples. Here a dual test sample is built by adding, deleting and replacing words in a test sample, or swapping two words in the sample, thereby resulting in a closely related test case. In theory, a model equipped with relevant commonsense should give consistent predictions on a pair of dual test cases. However, we find that none of the models are able to reach such consistency. Instead, the models are confused by the modification, tending to give the same predictions over a pair of dual samples despite they may have different gold labels. This further reveals that commonsense contained in the pre-trained models may remain in a surface level, without deep semantic comprehension. We publicly release our datasets, named commonsense ability tests (CATs), and the test script at GitHub. \footnote{https://github.com/XuhuiZhou/CATS}

\section{Tasks for Evaluating Commonsense}
Commonsense ability can be broadly divided to two categories. First, a model with commonsense ability should have basic knowledge about the world, for example, \textit{water always goes down}. Second, it should have the ability to reason over commonsense knowledge, such as \textit{water always goes down because there is gravity on the earth} and \textit{if you are injured, you should go to the hospital}. To comprehensively test different models' commonsense ability, we synthesize six challenging tasks by taking positive and negative samples from existing benchmarks, and further introduce a new task called Conjunction Acceptability (CA). 

We reframe all the tasks into sentence-scoring tasks by substitution or concatenation. For example, we create positive and negative samples by replacing a pronoun in the sentence of a WSC question with the candidates to obtain a test instance as Table \ref{example}. A model is asked to score the sentences and we pick the sentence with the highest score as its prediction in a test instance. Below we introduce the data sources and reframed tasks in detail (the correct answer is \textbf{bolded}). 

\begin{table}[t]
\centering
    \begin{tcolorbox}[fontupper=\small, fontlower=\small]
        {\bf Original:}\\
        \textit{Paul tried to call George on the phone, but he wasn't successful.} \\
        Who is he? \\ 
        Candidate: A. Paul (correct) B. George
        \tcblower
        {\bf Reframed:} \\
        \textit{A. Paul tried to call George on the phone, but Paul wasn't successful.} (Positive sample)\\
        \textit{B. Paul tried to call George on the phone, but George wasn't successful.} (Negative sample)
    \end{tcolorbox}
    \caption{Example of reframing a WSC question; Note that there can be additional negative samples.}
    \label{example}
\end{table}

\subsection{Sense Making (SM)}
Introduced by \citeauthor{wang-etal-2019-make} (2019), this task tests whether a model can differentiate sense-making and non-sense-making statements. Given a pair of statements (i.e a test instance), it requires the model to choose the more sensible statement. One example is: \textit{\textbf{I work 8 hours a day}} / \textit{I work 25 hours a day}. This task conforms to our evaluation schema without a change. More examples are shown in the SM section of Table \ref{table1}. The statements typically differ only in one key word which covers nouns, verbs, adjectives, and adverbs.

\subsection{Winograd Schema Challenge (WSC)}
The Winograd Schema Challenge (WSC) dataset \cite{Levesque:2012:WSC:3031843.3031909} consists 273 instances of the pronoun resolution problem. Each instance contains a sentence with a pronoun referring to one of nouns; the original question is to pick the correct noun. For our task, we transform the test as shown in Table \ref{example}. More examples are shown in the WSC section of Table \ref{table1}. WSC is recognized as one of the most difficult commonsense datasets.

\subsection{Conjunction Acceptability (CA)} 
 As stated by \citeauthor{lobue-yates-2011-types} (2011), logic-based commonsense knowledge is an important part of world knowledge in addition to content-based knowledge. We aim to probe a model's ability to understand the logic relations in the language by extracting 189 positive samples from the WSC dataset and replacing the conjunction manually with another conjunction to obtain a negative sample. We pair the positive and negative samples to obtain a test instance. For example, \textit{The lawyer asked the witness a question, and the witness was reluctant to answer it} / \textit{\textbf{The lawyer asked the witness a question, but the witness was reluctant to answer it}}. More examples are shown in the CA section of Table \ref{table1}.
 This task using ``because'', ``before'', ``when'', ``but'', ``and'' to correspond to the Cause and Effect, Preconditions, Simultaneous Conditions, Contradiction, and Addition logic relations, respectively. It is complementary to the other token-level tasks which focus more on content-based knowledge.

\subsection{SWAG}
SWAG \cite{zellers-etal-2018-swag} is a dataset with multiple choices questions about grounded situations. It questions models' understanding towards the relationship between two physical scenes. With the help of adversarial filtering (AF), \citeauthor{zellers-etal-2018-swag} created a sufficiently large amount of questions automatically. For example, given \textit{On stage, a woman takes a seat at the piano. She}, the question is to choose the following candidates: \textit{A. sits on a bench as her sister plays with the doll B. smiles with someone as the music plays C.is in the crowd, watching the dancers D. \textbf{nervously sets her fingers on the keys}}. We obtain a positive or negative sample by concatenating the context and a candidate together (e.g \textit{On stage, a woman takes a seat at the piano. She nervously sets her fingers on the keys}). There are one positive sample and three negative samples in a SWAG test instance. More examples are shown in the SWAG section of Table \ref{table1}. By forcing the model to predict the next action, it requires inductive reasoning and temporal reasoning. 

\subsection{HellaSwag}
HellaSwag \cite{zellers-etal-2019-hellaswag} is an argumented version of SWAG with the same data format as SWAG, more inference steps and higher data quality. While HellaSwag also includes the dataset from WikiHow, we choose only the instances coming from ActivityNet to make the results comparable to the original SWAG dataset. 

\subsection{Sense Making with Reasoning (SMR)}
Sense Making with Reasoning focuses on identifying the reason behind a statement \cite{wang-etal-2019-make} against commonsense. A model needs to understand that a specific statement (e.g \textit{can is usually made of gold}) is against commonsense and to make a choice for the reason behind from three candidates (e.g \textit{gold is too bright to make cans}, \textbf{\textit{gold is too soft to make cans}} and \textit{gold is too expensive to make cans}). We make a positive or negative sample by concatenating the statement and candidate reason together. For each test instance in SMR, there is a positive sample and two negative samples. More examples are shown in the SMR section of Table \ref{table1}. This task is intuitively difficult since it requires a model to have deeper knowledge of with higher-level inference, which belongs to abductive reasoning.

\subsection{Argument Reasoning Comprehension Task (ARCT)}
Similar to SMR, \citeauthor{habernal-etal-2018-argument} (2018) propose the ARCT dataset to test a model's abductive reasoning ability. 
Its domain lies in social topics such as search engine and LGBT rights, which is different from the daily-routine scenarios. For example, given a reason $R$: \textit{I find the idea that it is a sin to be born or live a life at all to be preposterous} and a claim $C$: \textit{Christians have created a harmful atmosphere for gays}, this task is to pick the correct warrant $W$ from two candidates: \textit{A. being gay isn't considered a sin B. \textbf{being gay is considered a sin}}, where $R \wedge W \rightarrow C$. We make a positive or negative sample by concatenating the reason, candidate warrant and claim together (e.g \textit{I find the idea that it is a sin to be born or live a life at all to be preposterous and since being gay is considered a sin, Christians have created a harmful atmosphere for gays}). A test instance in ARCT contains a pair of positive and negative samples. More examples are shown in the ARCT section of Table \ref{table1}. We further break this task into two variants, where ARCT1 represents the original dataset, ARCT2 represents an argumented dataset by adding negation to original instances to alleviate the statistical cues in the dataset \cite{niven-kao-2019-probing}.

We integrated the above test sets into a commonsense ability test (CATs) benchmark, released for future research.

\section{Pre-trained Models}
We take six contextualized representation models that give the state-of-the-art performances on NLP benchmarks such as GLUE \cite{wang-etal-2018-glue} and SQuAD \cite{rajpurkar-etal-2018-know}. Off-the-shelf models are taken. Below we give the detailed settings. 

\textbf{GPT} \cite{rad-2018} is a uni-directional transformer LM trained on 800M tokens of BookCorpus \cite{Zhu2015AligningBA}. 
 Given a text sequence $\mathbf{x} = [x_1, ..., x_T]$, GPT works in a way similar to conventional auto-regressive (AR) LM:
\[\max_{\theta} \log p_{\theta}(\mathbf{x}) = \sum_{t=1}^T\log p_{\theta}(x_t|\mathbf{x}_{<t}),\]
where $\mathbf{x}_{<t} = [x_1, ..., x_{t-1}]$. The model has dimension of hidden states $H = 768$, attention head numbers $A=12$, number of layers $L=12$ and total parameter size $P=110M$. 

\textbf{GPT2} \cite{radford2019language} works similarly as GPT with a few modifications on the hyperparameters. In particular, GPT2 optimizes the layer normalization, expands the vocabulary size to 50,257, increases the context size from 512 to 1024 tokens, and optimizes with a larger batchsize of 512. In addition, GPT2 is pre-trained on WebText, which was created from scraping web pages. 
The dataset roughly contains 8 million documents (40 GB). We study GPT2-base and GPT2-medium, with model size $H=768, A=12, L=12, P=117M$ and $H=1024, A=16, L=24, P=345M$, respectively, where the definitions of H, L and A are the same as for GPT.

\textbf{BERT} \cite{devlin-etal-2019-bert} jointly trains on a masked language modeling task and a next sentence prediction task (NSP). The model is trained on the BookCorpus and English Wikipedia, a total of approximately 3300M tokens. BERT is designed with the following objective:
\[\max_{\theta} \log p_{\theta}(\bar{\mathbf{x}}|\tilde{\mathbf{x}}) \approx \sum_{t=1}^T m_t\log p_{\theta}(x_t|\tilde{\mathbf{x}}) ,\]
where $\tilde{\mathbf{x}}$ is a corrupted version of text sequence $\mathbf{x}$, and $\bar{\mathbf{x}}$ is  masked tokens. $m_t=1$ if token $x_t$ belongs to $\bar{\mathbf{x}}$. 

Here we consider BERT-base and BERT-large, with $H=768, A=12, L=12, P=117M$ and $H=1024, A=16, L=24, P=340M$, respectively, where the definitions of H, L and A are the same as for GPT.

\textbf{XLNet} \cite{zhilin-19} is trained with a permutation-based language modeling objective to capture bidirectional contexts while retain the benefits of AR models. Specifically, they let $\mathcal{Z}_T$ be the set of all possible permutations of the length-T sequence $\mathbf{x} = [x_1, ..., x_T]$:
\[\max_{\theta} \mathbf{E}_{\mathbf{z} \sim \mathcal{Z}_T} \left[ \sum_{t=1}^T m_t\log p_{\theta}(x_{z_t}|\tilde{\mathbf{x}}_{\mathbf{z}<t}) \right] ,\]
where $z_t$ and $\mathbf{z}_{<t}$ are the $t$-th element and the first $t-1$ elements of a permutation $\mathbf{z} \in \mathcal{Z}_T$, respectively. In this way, XLNet ensures that any specific token $x_t$ in $\mathbf{x}$ has seen all the tokens before or after it.  

We consider XLNet-base and XLNet-large, whose model sizes are $H=768, A=12, L=12, P=117M$ and $H=1024, A=16, L=24, P=340M$, respectively, where the definitions of H, L and A are the same as for GPT. Note that XLNet-base is trained with the same data as BERT, while XLNet-large is trained with a larger dataset that consists of 32.98B subword pieces coming from Wiki, BookCorpus, Giga5, ClueWeb, and Common Crawl.

\textbf{RoBERTa} \cite{liu2019roberta} has the same architecture as BERT but is trained with dynamic masking, FULL-SENTENCES without NSP loss, a larger batch-size and a larger vocabulary size. Given the optimized design choice, one key difference of RoBERTa with other models is its large training dataset, which consists of BookCorpus, CC-NEWS, OpenWebText, and STORIES. With a total 160GB text, RoBERTa has access to more potential knowledge than the other models. 

\section{Experimental Design}
The CAT datasets are applicable to any model that has a method to score a sentence. They fit with the pre-trained models above, which are by nature language models. We derive the score of a sentence below with uni-directional-context LMs and bi-directional-context LMs, respectively. 

\begin{table*}[t]
\centering

\begin{tabular}{c | c c c | c c c c c | c}
    \toprule
     & CA & WSC & SM & SMR & SWAG & HellaSwag & ARCT1 & ARCT2 & Average \\ 
     \hline
    RANDOM &0.500& 0.500& 0.500& 0.333& 0.250& 0.250& 0.500& 0.500 & 0.416\\
    GPT &0.831& 0.555& 0.735& 0.318& 0.603& 0.261& 0.471& 0.524& 0.537\\
    GPT2-base &0.787& 0.512& 0.706& 0.355& 0.522& 0.300& 0.466& 0.509& 0.520 \\
    GPT2-medium	&0.885& 0.569& 0.746& 0.385& 0.597& 0.338& 0.462& 0.527& 0.564\\
    BERT-base &0.891& 0.576& 0.697& 0.418& 0.631& 0.364& 0.468& 0.503& 0.569\\
    BERT-large &0.934& 0.618& 0.694& 0.444& 0.683& 0.395& 0.477& 0.517& 0.595 \\
    XLNet-base &0.809& 0.544& 0.662& 0.374& 0.551& 0.381& 0.516& 0.526& 0.545 \\
    XLNet-large &0.891& 0.636& 0.583& 0.394& 0.692& 0.435& 0.563& 0.570& 0.596 \\
    RoBERTa-base &0.956& 0.625& 0.750& 0.423& 0.691& 0.414& 0.500& 0.537& 0.612\\
    RoBERTa-large &0.962& 0.693& 0.792& 0.512& 0.761& 0.500& 0.543& 0.599& 0.670\\
    \hline
    HUMAN &0.993 &0.920 &0.991 &0.975 &0.880 &0.945 &0.909 &0.909 & 0.945\\
    \toprule
\end{tabular}
\caption{Accuracy for each pre-trained contextualizer on each test set. The rightmost column shows the average of accuracy score of each model.}\smallskip
\label{table2}
\end{table*}

Formally, suppose the sentence S of n words $S = \{w_1,..., w_{k-1}, w_k, w_{k+1},...,w_n\}$. We define the score of a sentence as:
\[Score(S) = \frac{\sum_{k=1}^nlog(P_\theta(w_k|context_k)}{n} ,\] 
where the denominator $n$ is for alleviating the influence of the sentence length to models' prediction, especially in sentence-level tasks. For a uni-directional model, $context_k = S_{<k} \equiv \{w_1, ..., w_{k-1}\}$. The numerator becomes $\sum_{k=1}^nlog(P_\theta(w_k|S_{<k}))$, which is factorized from $\log(P_\theta(w_1,..., w_{k-1}, w_k, w_{k+1},...,w_n))$. This is essentially a LM. For a bi-directional model, the $context_k = S_{-k}$, which represents the $S$ with the $k$-th word being removed. In particular, the $k$-th word can be removed with being replaced by a special token `[MASK]' in BERT. The numerator $\sum_{k=1}^n \log(P_\theta(w_k|S_{-k}))$ can also be factorized from $\log(P_\theta(w_1,..., w_{k-1}, w_k, w_{k+1},...,w_n))$ under the assumption that $w_k$ is independent of the successive words (i.e. $w_{k+1}, w_{k+2}, ..., w_n$), which is the bi-directional-context LM.

Intuitively, $P_\theta(w_k|context_k)$ can be interpreted as how probable a word $w_k$ is given the $context_k$: $S_{<k}$ or $S_{-k}$. For example, let $S_{-k}=$ \textit{He put an [MASK] into the fridge}, $w_{k1} = elephant$ and $w_{k2} = turkey$. $P_\theta(w_{k2}|S_{-k})$ should have a relatively larger value since filling in the ``elephant'' in the first case results in an improper sentence, which is against commonsense.

As introduced earlier (Table \ref{table1}), all CATs tasks consist of instances with positive and negative sentences. After we score each sample in a test instance, the models predict the positive sample simply by taking the highest score in the instance.

\section{Commonsense Tests Results}

Table \ref{table2} shows the model performances with random choices as the baseline. Take WSC for example, the random baseline is 0.5, the human is 0.920 and all the models range between 0.512 and 0.693 with RoBERTa-large giving the best result of 0.693. Except for the ARCT task, all tested models demonstrate stronger performances than RANDOM, which indicates that the models all have varying degrees of commonsense. However, for most of the tasks, all of models are well below human performance. 

\subsection{Uni-directional Vs Bi-directional LM}
 We compare uni-directional (GPT, GPT2-base, GPT2-medium) and bi-directional models (Bert-base, Bert-large, XLNet-base, XLNet-large, RoBERTa-base, and RoBERTa-large). Picking the strongest model from each group, RoBERTa-large outperforms GPT2-medium by a large margin for every task. As mentioned before, RoBERTa-large has the same parameter size as GPT2-medium. However, RoBERTa-large is trained with much more data than GPT2-medium. 

\begin{figure}[t]
\centering
\includegraphics[width=0.9\columnwidth]{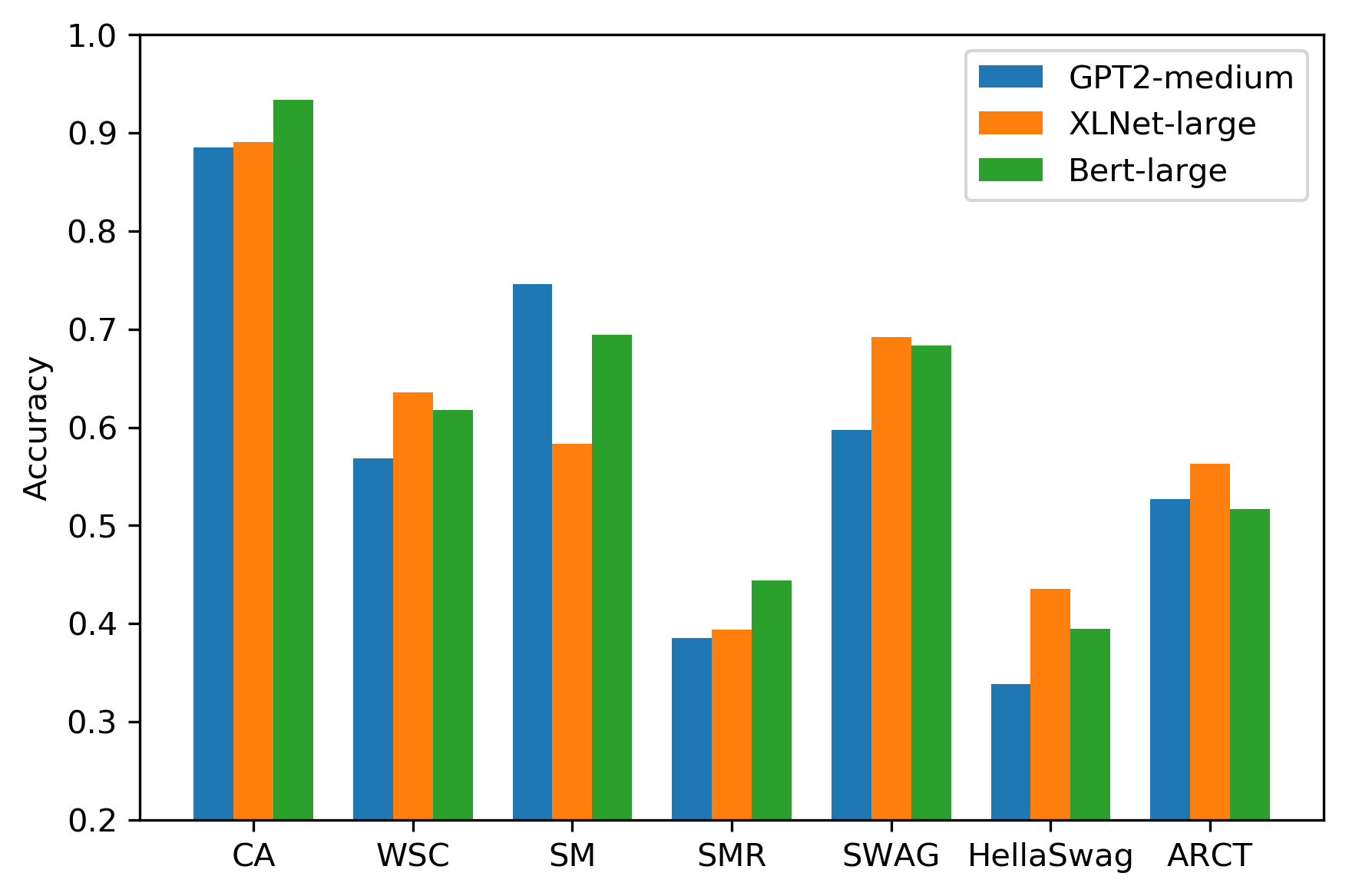} 
\caption{Comparison between bidirectional and unidirectional models among different tasks.}
\label{fig1}
\end{figure}
From Figure \ref{fig1}, we can see that except for the SM task, both BERT-large and XLNet-large outperform GPT2-medium while BERT-large is trained with a smaller dataset than GPT2-medium. This indicates that bi-directional context can be more useful for learning commonsense. Intuitively, the models with bi-directional context can make more sentence-level inference. 
While only the predecessing words receive sufficient context in a uni-directional model, every word has the full context for bi-directional models. Table \ref{example2} shows examples where RoBERTa-large makes the correct prediction but GPT2-medium does not, we can see that the key tokens, which are considered to be the most influential part in making the correct prediction, lie in the middle of the sentence. This can be the main reason why bi-directional context is important for models' commonsense ability.

\begin{table}[t]
\centering
    \begin{tcolorbox}[fontupper=\small, fontlower=\small]
        {\bf Token-level:}\\
        A. \textbf{Sam pulled up a chair to the piano, but the \textcolor{blue}{chair} was broken, so he had to stand instead.}\\
        B. Sam pulled up a chair to the piano, but the \textcolor{blue}{piano} was broken, so he had to stand instead.
        \tcblower
        {\bf Sentence-level:} \\
        A. \textbf{Comments sections permit a reader to analyze many different perspectives in one place, and since \textcolor{blue}{I want to see all these ideas}, even stupid ones, Comment sections have not failed.}\\
        B. Comments sections permit a reader to analyze many different perspectives in one place, but since \textcolor{blue}{I don't want to see all these stupid ideas}, Comment sections have not failed.
   
    \end{tcolorbox}
    \caption{Examples that a bi-directional model (RoBERTa-large) predicts correctly while a uni-directional model (GPT-2-medium) makes an incorrect prediction; the correct answer is \textbf{bolded}; the key tokens are colored in \textcolor{blue}{blue}.}
    \label{example2}
\end{table}

\subsection{Scale of Training Data}

A larger training dataset intuitively allows a model to have access to more commonsense knowledge, thus performs better in our tests. Trained with by far the most data, RoBERTa is the winner for every task. Most of the models are in fact trained on a subset of the dataset used to train RoBERTa. However, larger dataset do not always work when the model capacity is limited with regard to commonsense. For example, GPT2-base underperforms GPT for many tasks in our dataset, which suggests GPT2-base underfits the WebText dataset with regard to commonsense. The fact that RoBERTa-base has the same parameter size as GPT2-base, yet benefits from the larger dataset suggests that bi-directional models have larger representative power in commonsense ability.
\begin{figure}[t]
\centering
\includegraphics[width=0.9\columnwidth]{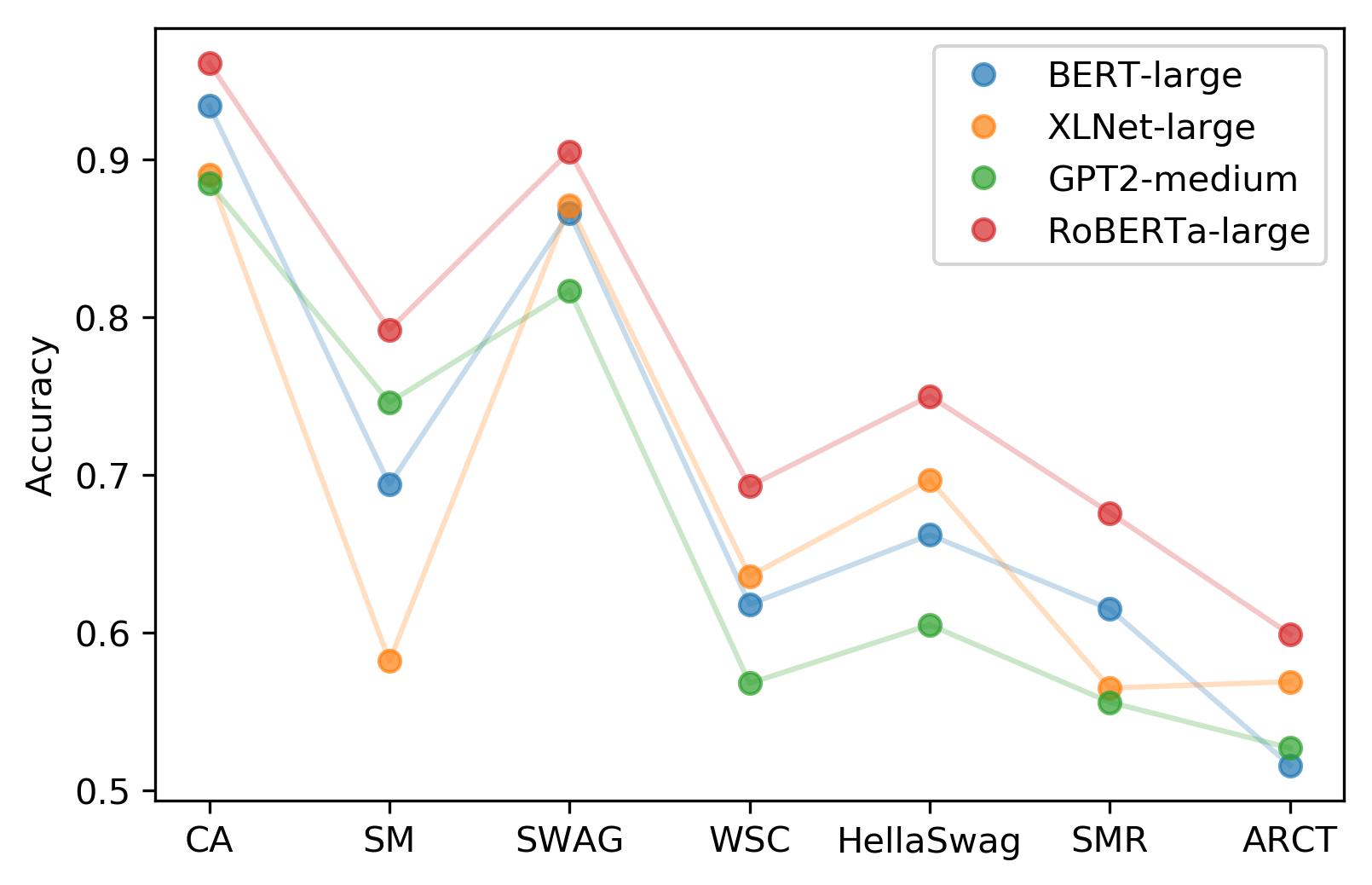} 
\caption{Models performances when the number of inference step (IS) increases. Tasks are ranked according to their IS in an increasing order from left to right. }
\label{fig2}
\end{figure}

\subsection{Number of Inference Steps}
Similar to humans, the model performance can intuitively drop when commonsense inference becomes more complicated. To verify this intuition, we pick 100 sentences randomly from each test dataset and annotate the number of required inference steps (IS) of each instance manually. The inference step of each test dataset is defined as the average of the number of the turns of reasoning necessary for the instances from the test dataset. We choose to answer the question 
by counting the logical operations that exist in an instance. For example, for the sentence 

\textit{They add a lot to the piece and I look forward to reading comments, but since comments sections always distract me from my work, Comment sections have failed.}, the logic chain is (\textit{They add a lot to the piece} $\wedge$ \textit{I look forward to reading comments}) $\wedge$ \textit{comments sections always distract me from my work} $\rightarrow$ \textit{Comment sections have failed}. Thus, this instance needs three inference steps.
 
In this way, we obtain the Inference Step (IS) for seven test datasets. Each instance is labeled by two expert annotators, and the inter-annotator agreement is 93\%. The final IS is the average from both annotators. Figure \ref{fig2} shows the results \footnote{The performances on tasks with more than one negative sample are transformed to binary-choice scales.} on the test cases with different IS. There is a decrease of performances as IS increases. SWAG and HellaSwag fall out the trend, which may suggest that the models have stronger commonsense ability in temporal reasoning.

Generally speaking, all of our tested models outperform the random baselines except for the ARCT task, which suggests that despite of using different modeling schemas, language modeling stands as an effective objective for extracting commonsense knowledge from large, raw texts. For each task, the overall performance increases with a larger model parameter size, a more sophisticated model design, and larger training data.

\section{Robustness Test}
The robustness of models in commonsense reasoning is an important perspective in evaluating deep commonsense ability. Intuitively, a person can reason whether a statement makes sense or not because he has consistent knowledge. If the statement changes slightly, for example, changing a key word, that person should still make the correct judgement.

\begin{figure}[t]
\centering
\includegraphics[width=1.0\columnwidth]{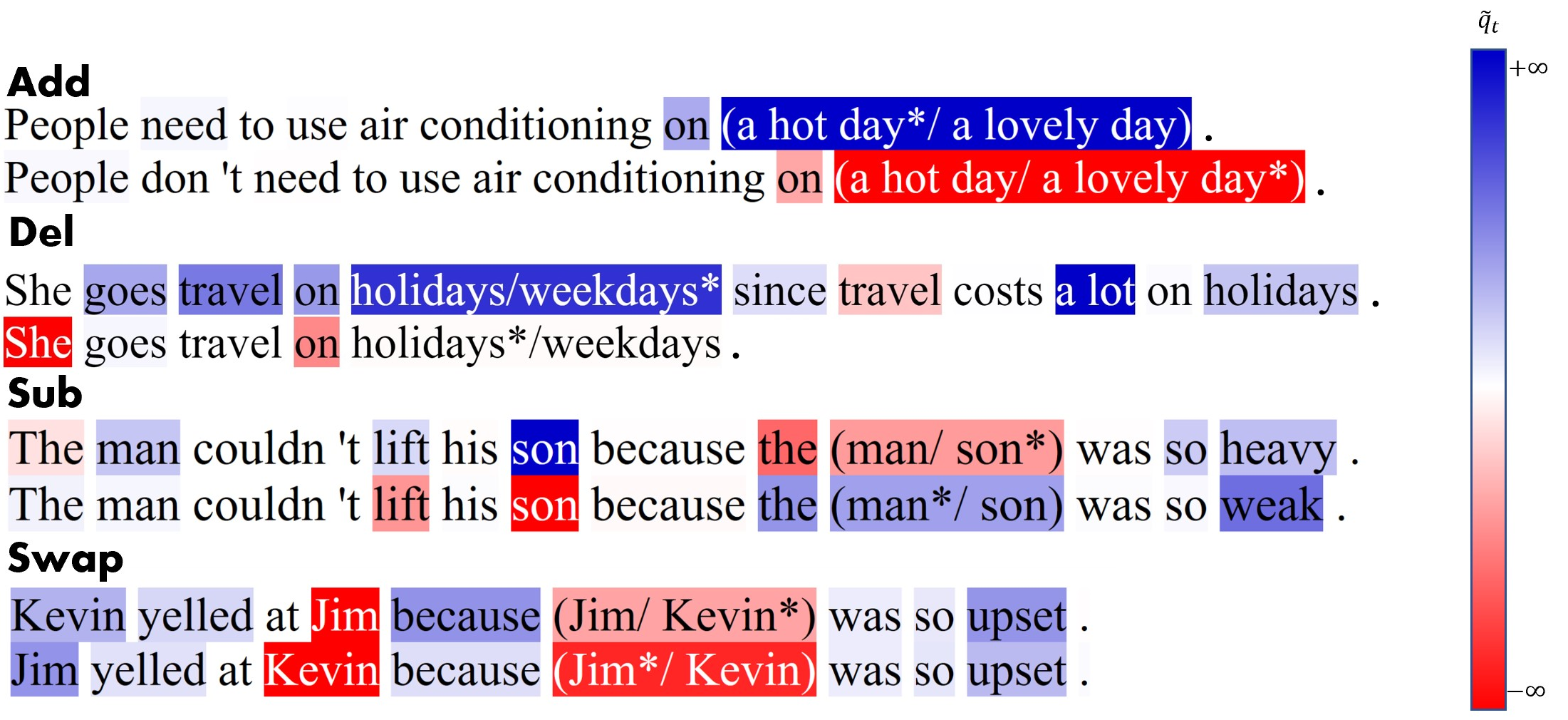} 
\caption{Samples of questions from Add, Del, Sub and Swap predicted correctly for the original instance but incorrectly for the dual instance. Note that a sentence here represents a test instance with a pair of positive and negative samples, represented by (.../...). Here we mark the correct prediction by an asterisk and display the normalized $q_t$ by coloring its corresponding word.}
\label{fig4}
\end{figure}

We aim to test the robustness of the five models by making dual test samples. A dual instance to the original instance should test the same commonsense knowledge point or largely relevant to the original one. In this way, we expect that the model can demonstrate consistency in the decision. One example is shown in Table \ref{example3}, which choosing A in the original instance should lead to choosing B in the dual case (See Figure \ref{fig4} for more examples). 

\begin{table}[t]
\centering
    \begin{tcolorbox}[fontupper=\small, fontlower=\small]
        {\bf Original:}\\
        A. People usually like wealth. B. People hardly like wealth.\\
        {\bf Dual:} \\
        A. People usually hate wealth. B. People hardly hate wealth.
    \end{tcolorbox}
    \caption{Example of a robust test case; it contains a test instance from the original test set with a dual test instance created manually. When the key word changes from `like' to `hate', the correct answer switches from A to B. This is an unique feature of our robustness test sets.}
    \label{example3}
\end{table}

We consider multiple ways to construct a dual test instance. Particularly, a dual test instance is built by methods: adding, deleting and replacing words in a test sample, or swapping two words in the sample, thereby resulting in a closely related test instance. All of our dual test instances are constructed from the original commonsense test data. 

We construct 75 dual instances for each method above over WSC, SM, and ARCT, keeping the instances from each dataset approximately equivalent in order to evaluate the influence of different duality methods to the models. We then pair each dual instance with the original instance to form a new test case. If the model gives the correct or wrong prediction for both of the instances in this case, we recognize it as a \textit{consistent} case.

\begin{table}[t]
\centering

\begin{tabular}{c c c c c}
\toprule
 & Add & Del & Swap & Sub \\ 
\hline
RANDOM &0.50& 0.50& 0.50& 0.50  \\
GPT &0.16& 0.22& 0.45& 0.23 \\
GPT2-base &0.20& 0.23& 0.46& 0.21 \\
GPT2-medium	&0.24& 0.24& 0.51& 0.24  \\
BERT-base &0.26& 0.15& 0.51& 0.29  \\
BERT-large &0.26& 0.26& 0.54& 0.27  \\
XLNet-base &0.16& 0.16& 0.41& 0.27  \\
XLNet-large &\textbf{0.36}& \textbf{0.39}& 0.36& 0.23  \\
RoBERTa-base &0.20& 0.27& 0.46& 0.36 \\
RoBERTa-large &0.29& 0.33& \textbf{0.55} & \textbf{0.44} \\
\toprule
\end{tabular}
\caption{Portion of consistent cases of each method for each contextualizer. Add stands for adding key words in the test sample; Del stands for deleting key words in the test sample; Swap stands for swapping the position of words in the test sample; Sub stands for replacing key words in the test sample. The best contextualizer for each method is \textbf{bolded}.}\smallskip \label{exp_dataset}
\label{table4}
\end{table}

The results are shown in Table \ref{table4}. In theory, a model equipped with relevant commonsense should give consistent predictions on a pair of dual test case. However, we find that none of the models reach consistency. In fact, their consistency is well below the random baselines except for the Swap method.

To better investigate the reason behind the poor consistency, we look at inconsistent cases from the pre-trained model (i.e RoBERTa-large). Similar to \citeauthor{Trinh-2018-a} (2018), we investigate how the model makes decision between two candidate sentences $S_{correct}$ and $S_{incorrect}$ where they have the same number of words. In particular, we look at:
\[ q_k = log(\frac{P_\theta(w_k|S_{correct}-\{w_k\})}{P_\theta(w_k|S_{incorrect}-\{w_k\})}),\]
where $1 \leq k \leq n$. It follows that the choice between $S_{correct}$ and $S_{incorrect}$ is made by the value $Q = \sum_k q_k$ being bigger than 0 or not. Visualizing the value of each $q_k$ provides more insights into the decisions of the model.

From Figure \ref{fig4}, we can tell that the model is confused by the modification, tending to give the same predictions over a pair of dual samples despite that they have different gold labels, especially for Sub, Add and Del. This further reveals that the commonsense knowledge contained in the pre-trained models may remain in a surface level, without deep semantic comprehension. 

\section{Related Work}

\citeauthor{liu-etal-2019-linguistic} (2019) evaluate BERT \cite{devlin-etal-2019-bert}, GPT \cite{rad-2018}, and ELMo \cite{peters-etal-2018-deep} on a variety of linguistics tasks. Their results suggest that the features generated by pre-trained contextualizer are sufficient for high performance on a board set of tasks but models fail on tasks requiring fine-grained linguistics knowledge. \citeauthor{Tenney2019WhatDY} (2019) evaluate similar models on a variety of sub-sentence linguistic analysis tasks. Their results suggest that contextualized word representation encode both syntax and semantics. Our work is in line in the sense that contextualized representation encode rich knowledge to be `probed'. However, we focus on evaluating the commonsense in those representations. To our best knowledge, this is the first work to systematically evaluate commonsense in pre-trained models. 

Our evaluation method is similar to \citeauthor{Trinh-2018-a} (2018), who make use of LM to score a sentence. However, they focus on Winograd schema questions with only self-trained recurrent LMs while we test five models' commonsense with seven diverse tasks.

\section{Conclusion}
We studied the commonsense knowledge and reasoning ability of pre-trained contextualizers with a suite of seven diverse probing tasks, showing that large-scale pre-trained contextualized representation has a certain degree of commonsense knowledge, but there is still a quite large gap between the current state-of-the-art representation models and robust human-level commonsense reasoning, which may require more breakthrough in modeling. We release our test sets, named CATs, publicly.

\section{Acknowledgments}
We would like to thank the anonymous reviewers
for their insightful comments, and Mr. Cunxiang Wang for his help on the collection of the data. Yue Zhang is the corresponding author.
\bibliographystyle{aaai}
\bibliography{aaaibib.bib}

\end{document}